\documentclass[5p,times,twocolumn]{elsarticle}

\usepackage{amsmath,amssymb}
\usepackage{booktabs}
\usepackage{graphicx}
\usepackage{subcaption}
\usepackage{makecell}
\usepackage{multirow}
\usepackage{float}
\usepackage{array}
\usepackage{tabularx}
\usepackage{xcolor}
\usepackage{hyperref}
\usepackage{natbib}
\usepackage{microtype}   

\definecolor{myblue}{HTML}{2166AC}
\definecolor{myred}{HTML}{D6604D}
\definecolor{mygreen}{HTML}{1A9641}
\definecolor{mygray}{HTML}{6B6B6B}

\graphicspath{{figures/}}

\journal{Artificial Intelligence in Medicine}

\begin{document}

\begin{frontmatter}

\title{From Latent Biomarkers to Clinical Rules: Embedding-Guided
       Rule Mining and Attribution-Based Translation for
       Interpretable Tabular Learning}

\author[]{Majid Lotfian Delouee\corref{cor1}}
\author[]{Hamed Ayoobi\corref{cor1}}
\author[]{Sjors G. J. G. In ’t Veld}
\author[]{Martijn C. Schut}

\cortext[cor1]{Corresponding authors}


\begin{abstract}

Clinical decision support tools are most useful when they can provide not just accurate predictions but also the reasoning behind them. Rule-based models offer this transparency, but rules derived directly from raw clinical measurements tend to miss patterns that emerge from multi-variable interactions. This paper presents a four-step pipeline that addresses this limitation. A Feature Tokenizer and Transformer (FT-Transformer) first learns a compact patient representation from raw tabular features. Embedding dimensions that consistently separate patient groups are then identified as latent biomarkers: data-derived proxies of disease state. Decision rules are mined in this latent space using small decision trees, and each selected rule is translated back into measurable clinical features through combined gradient-input saliency and CLS attention attribution. The complete framework was evaluated on six publicly available clinical and population health datasets spanning neurometabolic, cardiac, and metabolic conditions, with sample sizes ranging from 303 to 253,680 records. Rules were extracted at four embedding dimensions (16, 32, 64, and 128) and compared against rules mined directly from raw features using the area under the receiver operating characteristic curve (AUROC), precision, recall, F1 score, and coverage on a held-out test set. Translated rules outperformed raw-feature rules in five of the six datasets,
with mean AUROC gains ranging from 0.04 on the Cleveland Heart Disease dataset to 0.23 on the CDC Diabetes Health Indicators dataset. The heart disease dataset was the exception: embedding-space rules achieved an AUROC of 0.98, but the translation step reduced this to 0.72, reflecting
the difficulty of compressing a high-fidelity latent rule into two conditions on raw features.
Routing rule discovery through the transformer's latent space, and then attributing those rules back to clinically measurable features, offers a practical path toward decision support tools that are both predictively strong and expressed in language that clinicians can evaluate and act on.

\end{abstract}


\end{frontmatter}


\section{Introduction}
\label{sec:introduction}

Clinicians rarely make diagnostic decisions based on a single measurement.
The risk of hospital readmission in
a diabetic patient, or the likelihood of a cardiac event all depend on reading
multiple clinical indicators together, each individually incomplete.
Clinical decision support systems are designed to assist this process by
providing structured guidance derived from patient data, and the format that
physicians most readily adopt is the explicit IF-THEN rule.
A statement such as ``if fasting glucose exceeds 126~mg/dL and BMI is above
30~kg/m\textsuperscript{2}, flag for diabetes screening'' is transparent,
auditable, and can be embedded directly into an electronic health record (EHR)
workflow without requiring clinicians to treat the system as a black box.

The standard way to generate such rules is to train a decision tree directly
on measured feature values.
This is reliable and well understood, but it carries a structural limitation:
each split acts on a single feature at a time, and the features are used
exactly as they appear in the patient record.
Patterns that require multiple variables to interact, or that are partly
obscured by measurement noise and scale differences, are difficult to capture
by greedy splits on raw values.
Rules built this way often show acceptable performance on the training
population but generalize poorly, particularly in datasets where class
imbalance, missing values, or many weakly predictive features are present.

Transformer architectures adapted for tabular inputs have substantially
changed what is achievable in this space.
The Feature Tokenizer and Transformer~\citep{gorishniy2021} projects each
clinical measurement, whether a continuous laboratory value, a categorical
symptom grade, or a binary indicator, into a shared token representation and
processes all features jointly through multi-head self-attention.
The resulting patient representation, a compact vector in a learned latent
space, encodes feature interactions that individual measurements do not reveal
on their own.
On tabular benchmarks covering a range of clinical domains, this architecture
matches or surpasses gradient-boosted tree ensembles when sufficient training
data are available~\citep{grinsztajn2022,shwartzziv2022}.

\begin{figure*}[t]
  \centering
  \includegraphics[width=\linewidth]{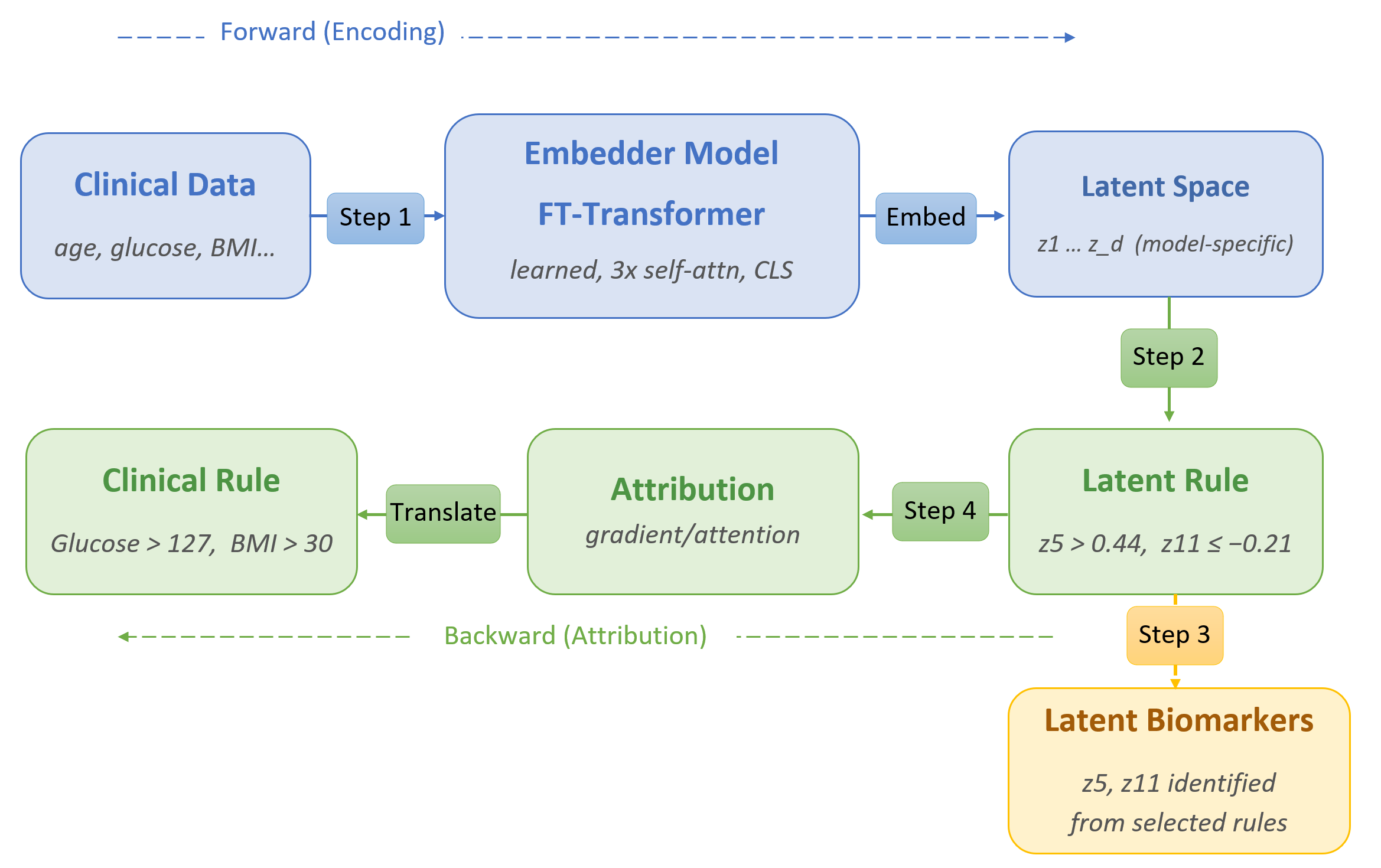}
  \caption{Conceptual overview of the proposed pipeline. Clinical patient
  records are encoded into a compact latent space by a trained FT-Transformer
  (forward path, blue). Interpretable rules are mined in the latent space and
  the embedding dimensions involved are identified as latent biomarkers.
  Gradient and attention attribution then maps each rule back to raw clinical
  features, producing a translated rule in clinical language (attribution
  path, red).}
  \label{fig:concept}
\end{figure*}

The latent representation itself is not interpretable.
A patient described as a 32-dimensional vector, or a rule that says
``$z_3 > 0.52$ predicts high disease severity,'' conveys nothing that
a clinician can use or evaluate.
Deploying such a rule directly in a hospital would require unconditional trust
in a system whose reasoning is invisible, which conflicts with patient safety
requirements and the ethical standards of evidence-based medicine
\citep{rudin2019,ghassemi2021}.

The work presented here bridges this gap.
We propose treating the embedding dimensions that most strongly separate
patient groups as \textit{latent biomarkers}: hidden surrogates of disease
state that the model has learned to extract from the raw clinical variables.
In the same way that serum creatinine serves as a surrogate for kidney
function without directly measuring glomerular filtration rate, a latent
biomarker is a derived indicator whose discriminative power reflects an
underlying disease process captured by the model.
Decision rules are mined in this latent biomarker space, where class-relevant
structure is cleaner and more compact than in raw-feature space.
Each selected rule is then attributed back to raw clinical features using
gradient saliency and CLS attention weights from the transformer itself,
producing a translated rule that is again written in terms that appear in
a patient's chart.
Figure\ref{fig:concept} illustrates the overall approach.

This paper makes three contributions.
First, we formalize the concept of latent biomarkers for tabular clinical data:
embedding dimensions are ranked by their AUROC when used as a one-dimensional
predictor, and those that pass a confidence and coverage threshold in the
extracted rules are designated as biomarkers for the corresponding outcome
class.
Second, we describe a combined attribution method that translates
embedding-space rules into raw-feature conditions by aggregating
gradient-input saliency across all latent dimensions in a rule and weighting
it with CLS attention scores, then fitting a shallow decision tree on the
top-ranked raw features.
Third, we evaluate the complete framework on six publicly available clinical
datasets spanning neurometabolic, cardiac, and metabolic disease domains, and
demonstrate that translated rules outperform raw-feature rules in five of
six cases, with AUROC gains reaching 0.22 in the largest datasets.



\section{Related Work}
\label{sec:related}

\subsection{Deep Learning for Tabular Clinical Data}

Tabular machine learning in clinical settings has been dominated for many years
by gradient-boosted decision trees, owing to their tolerance of mixed feature
types, missing data, and small sample sizes.
Several rigorous benchmarks have confirmed that tree ensembles remain
competitive with deep learning on datasets with a few thousand samples or
fewer~\citep{grinsztajn2022}, but the advantage narrows as data volume grows
and as feature interactions become more complex.

Among deep learning approaches, attention-based architectures have attracted
particular interest because the self-attention mechanism provides a natural
model of how features co-determine each other's relevance.
TabTransformer~\citep{huang2020} applies contextual embedding only to
categorical features, leaving numeric variables as raw inputs to a final
multilayer perceptron.
TabNet~\citep{arik2021} uses a sequential attention mechanism that selects a
sparse subset of informative features at each prediction step, mimicking
feature selection but operating end-to-end with gradient descent.
The FT-Transformer~\citep{gorishniy2021} takes the most uniform approach:
every feature, categorical or numeric, is projected into a shared token
dimension, and the full token sequence is processed by a standard transformer
encoder augmented with a learnable CLS token.
The CLS token's output summarizes the patient record into a compact
representation that is passed to a classification head.
Across multiple tabular benchmarks, this architecture produces competitive
performance while also yielding representations amenable to downstream
analysis.

In clinical applications, transformer models have been used to model
longitudinal EHR sequences~\citep{rasmy2021}, to predict mortality in
intensive care units from time-series data, and to integrate structured and
unstructured clinical information.
The specific question of how to extract interpretable rules from
tabular transformer models, and how to translate those rules back into
clinical language, has received limited attention.
Our work addresses this gap directly.

\subsection{Interpretable Machine Learning in Medicine}

Interpretability has long been recognized as a prerequisite for clinical
adoption of machine learning tools.
\citet{rudin2019} argued compellingly that for high-stakes decisions, inherently
interpretable models should be preferred over post-hoc explanations of opaque
ones, because the explanation is always approximate and may not faithfully
represent the model's actual reasoning.
In clinical practice this concern is amplified: a mistaken explanation of a
cardiac risk score could lead to the wrong treatment, and the error may go
undetected precisely because the explanation seemed plausible.

Post-hoc explanation methods are widely used in clinical ML despite these
concerns.
SHAP values~\citep{lundberg2017} assign each feature a contribution to a
specific prediction based on cooperative game theory, and they have been
applied across a range of clinical prediction tasks including sepsis
detection, readmission risk, and cancer diagnosis.
LIME~\citep{ribeiro2016} fits a local linear surrogate model around each
prediction, offering a different perspective on feature importance.
Both methods explain individual predictions rather than generating global
decision rules, which limits their utility in workflows that require
consistent, auditable logic across many patients.

Global rule extraction offers a way to satisfy both the predictive demands
of modern machine learning and the auditability requirements of clinical
governance~\citep{holzinger2019}.
Several older approaches extract rules directly from trained feedforward
networks by discretizing hidden-unit activations or by using a rule learner
to approximate the network's input-output mapping~\citep{quinlan1993}.
More recent work has examined whether attention weights in transformer models
serve as reliable explanations of predictions~\citep{jain2019,serrano2019},
with mixed conclusions: attention captures some aspects of feature importance
but is not sufficient on its own.
The approach taken here uses attention as one of two complementary attribution
signals rather than as a primary explanation, addressing the known limitations
of attention-only interpretability.

The broader challenge of deploying interpretable AI in clinical settings has
been examined from regulatory and ethical perspectives~\citep{rajpurkar2022}.
\citet{ghassemi2021} cautioned that explanations of high-stakes clinical models
may create false confidence in both users and regulators if they do not
faithfully capture the model's decision logic.
The translated rules produced by our pipeline are not post-hoc approximations
of a black box: they are independently trained decision trees whose features
and thresholds are chosen based on the model's internal representation, and
their performance is measured separately on held-out data.

\subsection{Attribution Methods for Tabular Models}

Gradient-based attribution assigns feature importance by measuring how
sensitively the model output responds to small changes in each input,
weighted by the input magnitude~\citep{simonyan2014}.
For a continuous feature $x_m$ and a scalar output $f$, the gradient-input
saliency is $s_m = \left|\partial f / \partial x_m\right| \cdot |x_m|$,
which combines the model's local sensitivity with the scale of the feature
value.
Integrated gradients~\citep{sundararajan2017} extend this idea by averaging
gradients along a straight line from a reference input to the actual input,
satisfying an axiomatic completeness property.
In the context of our pipeline, gradient attribution is computed not with
respect to the final prediction but with respect to specific latent dimensions,
allowing us to identify which raw features drive the embedding coordinates
involved in a selected rule.

Attention-based attribution uses the CLS-to-feature attention weights produced
by each encoder layer.
Because the CLS token aggregates information from all feature tokens through
the attention mechanism, the weight assigned to each feature token reflects
its contribution to the final representation.
These weights are defined for all feature types, including categorical ones
for which gradient attribution is not directly applicable.
Averaging attention weights over the encoder layers and over rule-positive
test samples provides a stable, feature-level importance estimate.

Combining gradient and attention attribution has practical advantages for
tabular clinical data.
Gradient attribution is precise for numeric features but undefined for
categorical inputs; attention attribution covers all features but can be
noisy for individual samples.
Normalizing each signal independently and averaging them yields a unified
importance ranking that leverages the strengths of both approaches.
This combined strategy is the basis of the translation step described in
Section~\ref{sec:methods}.


\section{Materials and Methods}
\label{sec:methods}

\subsection{Datasets}
\label{sec:datasets}

\begin{table*}[ht]
\caption{Summary of the six clinical and population health datasets used in
this study. Sample sizes reflect the full dataset before splitting.
Class balance is reported as the percentage of the majority class.
Feature counts exclude the target column.}
\label{tab:datasets}
\centering
\small
\renewcommand{\arraystretch}{1.25}
\begin{tabularx}{\textwidth}{%
  l              
  l              
  r              
  r              
  r              
  r              
  X              
}
\toprule
\textbf{Dataset} &
\textbf{Domain} &
\textbf{Samples} &
\textbf{Features} &
\textbf{Classes} &
\textbf{Maj.\,\%} &
\textbf{Notes} \\
\midrule
Pima Indians Diabetes &
  Endocrinology &
  768 & 8 & 2 & 65\% &
  Female Pima Indian patients aged $\geq$21 years; binary diabetic outcome \citep{smith1988} \\[2pt]

Cleveland Heart Disease &
  Cardiology &
  303 & 13 & 2 & 54\% &
  UCI repository; presence/absence of coronary artery disease~\citep{uci_heart} \\[2pt]

Adrenoleukodystrophy (ALD) &
  \makecell{Neurometabolic disease/\\Lipidomics} &
  184 & 1811 & 5 & 58.2\% &
  Lipidomics dataset for five-stage ALD severity  \citep{jaspers2024lipidomic}\\[2pt]

Adult Census Income &
  Population Health &
  48{,}842 & 14 & 2 & 76\% &
  Annual income ${\leq}$50K / ${>}$50K; used as a proxy for
  socioeconomic health determinants~\citep{uci_adult} \\[2pt]

CDC Diabetes Health Indicators &
  Endocrinology &
  253{,}680 & 21 & 2 & 86\% &
  BRFSS 2015 telephone survey; binary diabetes outcome~\citep{cdc2023} \\[2pt]

Diabetes 130-US Hospitals &
  Endocrinology &
  101{,}766 & 47 & 2 & 89\% &
  Hospital admissions 1999–2008; outcome is 30-day readmission
  \citep{strack2014} \\
\bottomrule
\end{tabularx}
\medskip
\newline
\footnotesize
Maj.\% = percentage of samples belonging to the majority class.
All splits were stratified by class label.
\end{table*}

Six publicly available datasets were used (Table~\ref{tab:datasets}).
Three were selected as smaller benchmarks to test the pipeline under
limited-data conditions: the Pima Indians Diabetes Dataset~\citep{smith1988},
the Cleveland Heart Disease dataset from the UCI Machine Learning
Repository~\citep{uci_heart}, and anAdrenoleukodystrophy (ALD) severity
dataset with five ordered classes \citep{jaspers2024lipidomic}.
Three larger datasets were added to assess scalability: the Adult Census
Income dataset~\citep{uci_adult}, the CDC Diabetes Health Indicators dataset
derived from the Behavioral Risk Factor Surveillance System
survey~\citep{cdc2023}, and the Diabetes 130-US Hospitals dataset covering
a decade of inpatient diabetic admissions~\citep{strack2014}.

The adrenoleukodystrophy (ALD) dataset was derived from a plasma lipidomics study investigating associations between lipid profiles and disease severity in X-linked ALD \citep{jaspers2024lipidomic}.
All other datasets involve binary classification.
For comparability, multi-class rules in the ALD dataset were evaluated per class using a one-versus-rest formulation.

\subsection{Preprocessing and Data Splitting}
\label{sec:preprocessing}

All preprocessing followed a strict leakage-safe protocol.
Each dataset was partitioned into training (60\%), validation (20\%), and
test (20\%) subsets using stratified random sampling with a fixed random seed
(42), so that class proportions were preserved across splits.
The test set was held out entirely and never used for model selection,
hyperparameter tuning, or preprocessing estimation.

Numeric features were imputed with the training-set median and then
standardized using the training-set mean and standard deviation.
Categorical features were integer-encoded with a vocabulary built from the
training set; values unseen during training were mapped to a reserved
unknown token.
No dataset-level normalization or feature engineering was performed beyond
these steps, so the pipeline could be applied uniformly to all six datasets
without domain-specific customization.

\subsection{Step 1: Learning Patient Embeddings}
\label{sec:step1}

An FT-Transformer was trained independently for each dataset and each
embedding dimension $d_r \in \{16, 32, 64, 128\}$, yielding four models
per dataset.
The architecture follows~\citet{gorishniy2021}: each feature is independently
projected into a token of dimension $d_\text{token}$ (set to 64 for
$d_r \geq 32$, and 32 for smaller configurations) using a learned linear
layer for numeric features and an embedding lookup for categorical features.
A learnable classification (CLS) token is prepended to the feature token
sequence.
Three transformer encoder layers, each with multi-head self-attention
(8 heads for $d_\text{token} = 64$, 4 heads for $d_\text{token} = 32$),
GELU activations, and pre-layer normalization, process the complete token
sequence.

The CLS token output at the final encoder layer is passed through a two-layer
projection head with a GELU activation to produce the patient representation
vector $\mathbf{z} \in \mathbb{R}^{d_r}$.
A linear layer maps $\mathbf{z}$ to the class logits.
Dropout of 0.15 was applied throughout, and gradient norm clipping at 1.0
was used during training.

Training used the AdamW optimizer with a learning rate of $10^{-3}$ and
weight decay of $10^{-4}$.
Datasets with class imbalance used inverse-frequency class weights in the
cross-entropy loss.
Early stopping monitored macro-averaged F1 on the validation set, with
patience of 25 epochs and a maximum of 300 epochs.
The best-performing checkpoint was restored before embedding extraction.
After training, embeddings were extracted for all three splits by a single
forward pass through the frozen model.

\subsection{Step 2: Rule Mining in the Latent Space}
\label{sec:step2}


\begin{table*}[ht]
\caption{Mean test-set AUROC for the three rule types across all six datasets,
averaged over selected rules and embedding dimensions (16, 32, 64, 128).
The best value per dataset is in \textbf{bold}.
Delta shows the AUROC gain of the translated rule over the raw-feature rule.
A positive delta indicates that routing through the latent space and
translating back to raw features improved performance.}
\label{tab:auc_main}
\centering
\small
\renewcommand{\arraystretch}{1.3}
\begin{tabular}{l r r r r}
\toprule
\textbf{Dataset} &
\textbf{Raw-Feature Rule} &
\textbf{Embedding Rule} &
\textbf{Translated Rule} &
\textbf{$\Delta$ AUROC} \\
\midrule
Pima Indians Diabetes     & 0.623 & 0.610 & \textbf{0.779} & +0.156 \\
Cleveland Heart Disease   & 0.675 & \textbf{0.985} & 0.717        & +0.042 \\
Adrenoleukodystrophy (ALD)   & \textbf{0.986}$^\dagger$ & 0.863 & 0.862 & $-$0.124 \\
Adult Census Income       & 0.618 & 0.589 & \textbf{0.734} & +0.116 \\
CDC Diabetes Health Ind.  & 0.532 & 0.549 & \textbf{0.758} & +0.226 \\
Diabetes 130-US Hospitals & 0.537 & 0.526 & \textbf{0.753} & +0.216 \\
\midrule
\textit{Mean (5 datasets excl.\ Heart)} &
  \textit{0.659} & \textit{0.627} & \textit{0.777} & \textit{+0.118} \\
\bottomrule
\end{tabular}
\medskip
\newline
\footnotesize
AUROC = area under the receiver operating characteristic curve, computed as
$(\text{sensitivity} + \text{specificity})/2$ of the binary rule
(fires / does not fire) against ground-truth class labels on the test set.
$\Delta$ AUROC = Translated Rule AUROC $-$ Raw-Feature Rule AUROC.
$^\dagger$ALD raw mean computed over the 8 rules for which a raw-feature rule
was found; two classes (severity 2 and 4) had no raw-feature rule and are
excluded from the raw and embedding means but included in the translated mean.
\end{table*}

A decision tree classifier was trained on the training-set embeddings to
predict the clinical outcome.
Hyperparameters (maximum depth, minimum samples per leaf, maximum number
of leaves, cost-complexity pruning coefficient) were selected by grid search,
evaluating each candidate configuration on the validation-set AUROC.
The search grid covered depths of 2 through 6 (plus unconstrained),
minimum leaf sizes of 1, 2, 5, and 10, maximum leaf counts of 8, 16, 32,
and unconstrained, and pruning coefficients of $0$, $10^{-4}$, $10^{-3}$,
and $10^{-2}$.
Only leaf nodes covering at least 1\% of training samples were retained as
rules, to ensure that each rule applies to a clinically meaningful fraction
of patients.

The identical procedure was applied to the raw feature matrix, yielding
raw-feature rules as the comparison baseline.
Both the embedding tree and the raw-feature tree were tuned on the same
validation set, so any AUROC difference on the test set reflects the
informativeness of the latent representation rather than any tuning advantage.
Each embedding rule takes the form:
IF $z_{j_1}$ op$_1$ $\theta_1$ AND $z_{j_2}$ op$_2$ $\theta_2 \ldots$
THEN class $c$,
where $z_j$ denotes the $j$-th latent dimension, op is either $\leq$ or $>$,
and $\theta$ is a learned threshold.

\subsection{Step 3: Latent Biomarker Identification}
\label{sec:step3}

From the embedding rules produced in Step~2, up to $K = 3$ rules per class
were selected using a composite score equal to the training-set confidence
(fraction of leaf samples belonging to the predicted class) multiplied by
the training-set coverage (fraction of training samples that reach the leaf).
Only rules with confidence above 0.6 and coverage above 0.05 were eligible.
The latent dimensions $z_j$ appearing in the conditions of a selected rule
are the latent biomarkers for the corresponding class: they are the
coordinates of the embedding space that the model has learned to associate
with that clinical outcome.

Each selected rule was applied to the test-set embeddings to compute
precision, recall, F1 score, and AUROC on the held-out samples.
The indices of test samples for which the rule fires (rule-positive samples)
were recorded for use in Step~4.

\subsection{Step 4: Attribution and Rule Translation}
\label{sec:step4}

For each selected embedding rule, two attribution signals were computed
on the rule-positive test samples.
Figure~\ref{fig:concept} in the Introduction summarises the overall data flow;
the subsections below describe each step in detail.

\paragraph{Gradient-input saliency.}
For a rule involving latent dimensions $\mathcal{J} = \{j_1, j_2, \ldots\}$,
the saliency of numeric feature $x_m$ with respect to dimension $z_j$ is:
\begin{equation}
  s_{mj} = \left| \frac{\partial z_j}{\partial x_m} \right| \cdot |x_m|,
  \label{eq:saliency}
\end{equation}
where the partial derivative is obtained by a single backward pass through the
frozen model.
Saliency scores are averaged equally across all dimensions in the rule:
\begin{equation}
  \bar{s}_m = \frac{1}{|\mathcal{J}|} \sum_{j \in \mathcal{J}} s_{mj}.
  \label{eq:avg_saliency}
\end{equation}

\paragraph{Attention attribution.}
The CLS-to-feature attention weights from all three encoder layers were
extracted for each rule-positive test sample using the \texttt{need\_weights}
option of the PyTorch multi-head attention module~\citep{paszke2019}.
The attention weight from the CLS token to feature token $f$, averaged over
layers and over rule-positive samples, provides a mean attribution score for
each feature including categorical ones.

\paragraph{Combined score and translated rule.}
Gradient scores (numeric features only) and attention scores (all features)
were each normalized to sum to one within their respective feature sets and
then averaged to produce a combined importance ranking.
The top five features by combined score were used to fit a depth-2 decision
tree on the training set.
The binary target for this tree was whether a training sample falls in the
rule-positive region of the embedding space (as defined by the embedding rule).
This tree was fitted using scikit-learn~\citep{scikit} with balanced class
weights.
The leaf with the highest class-1 confidence defined the translated rule,
expressed entirely as conditions on raw clinical measurements.

\begin{figure*}[ht]
  \centering
  \includegraphics[width=\linewidth]{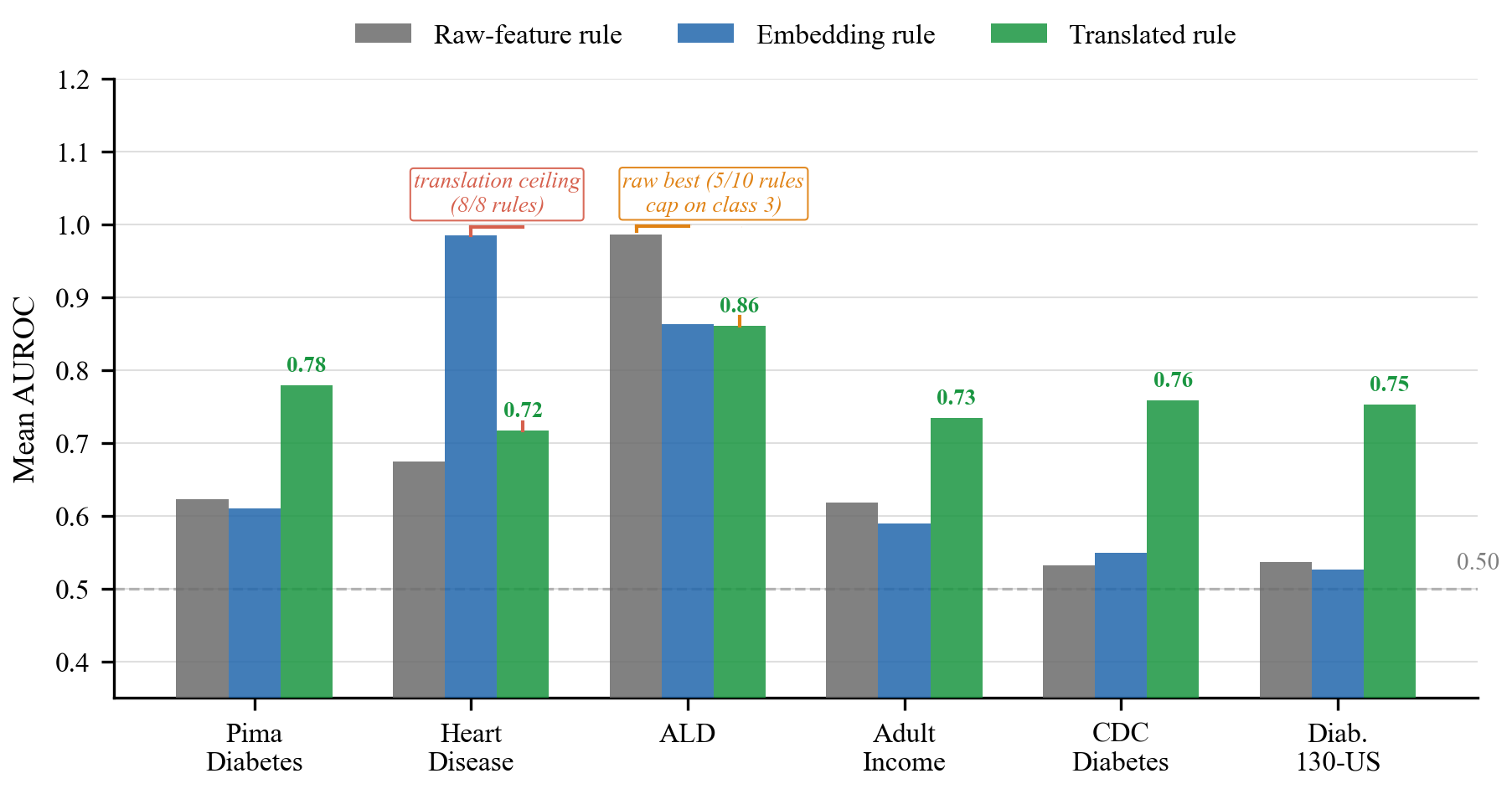}
  \caption{Mean test-set AUROC per dataset for raw-feature rules (gray),
  embedding-space rules (blue), and translated rules (green), averaged over
  all selected rules and embedding dimensions within each dataset.
  Translated rules outperform raw-feature rules in five of six datasets and achieve the highest AUROC among the three rule types in four of six datasets.
  Two exceptions are annotated: Heart Disease, where near-perfect embedding
  separation (mean Emb\,AUROC $= 0.99$) creates a translation ceiling
  (8/8 rules, Emb $\gg$ Trans); and ALD, where direct lipid biomarkers
  yield raw-rule AUROC near 1.0, a baseline translated rules cannot exceed
  (5/10 rules).}
  \label{fig:auc_bars}
\end{figure*}

\subsection{Evaluation Protocol}
\label{sec:eval}

Performance was assessed on the held-out test set using AUROC as the primary
metric, computed as balanced accuracy
$(\text{sensitivity} + \text{specificity}) / 2$ of the binary rule
(fires versus does not fire) against the ground-truth class label.
Secondary metrics were precision (fraction of rule-positive samples belonging
to the predicted class), recall (fraction of true-class samples captured by
the rule), F1 score, and coverage (fraction of all test samples for which the
rule fires).

All metrics were averaged over the selected rules within each dataset and
embedding dimension, then further averaged over the four embedding dimensions
to produce the per-dataset summary values reported in Section~\ref{sec:results}.
Statistical comparisons between translated and raw-feature rules were not
carried out formally due to the small number of datasets; results are reported
descriptively with the direction and magnitude of differences indicated.


\section{Results}
\label{sec:results}

\subsection{Overall Performance Comparison}
\label{sec:res_overall}

Table~\ref{tab:auc_main} summarizes the mean test-set AUROC averaged over all selected rules and embedding dimensions for each of the six datasets.
Translated rules outperformed raw-feature rules in five of the six datasets, with gains ranging from $+0.04$ (Cleveland Heart Disease) to $+0.23$ (CDC
Diabetes Health Indicators). 
The ALD dataset is the exception: raw-feature rules achieve a mean AUROC of $0.99$ across the eight raw-feature rules that were identified, driven in part by highly discriminative lipid features such as $TG 51:2$, setting a strong baseline that translated rules ($\Delta = -0.12$) do not exceed.

\begin{figure}[H]
  \centering
  \includegraphics[width=0.85\linewidth]{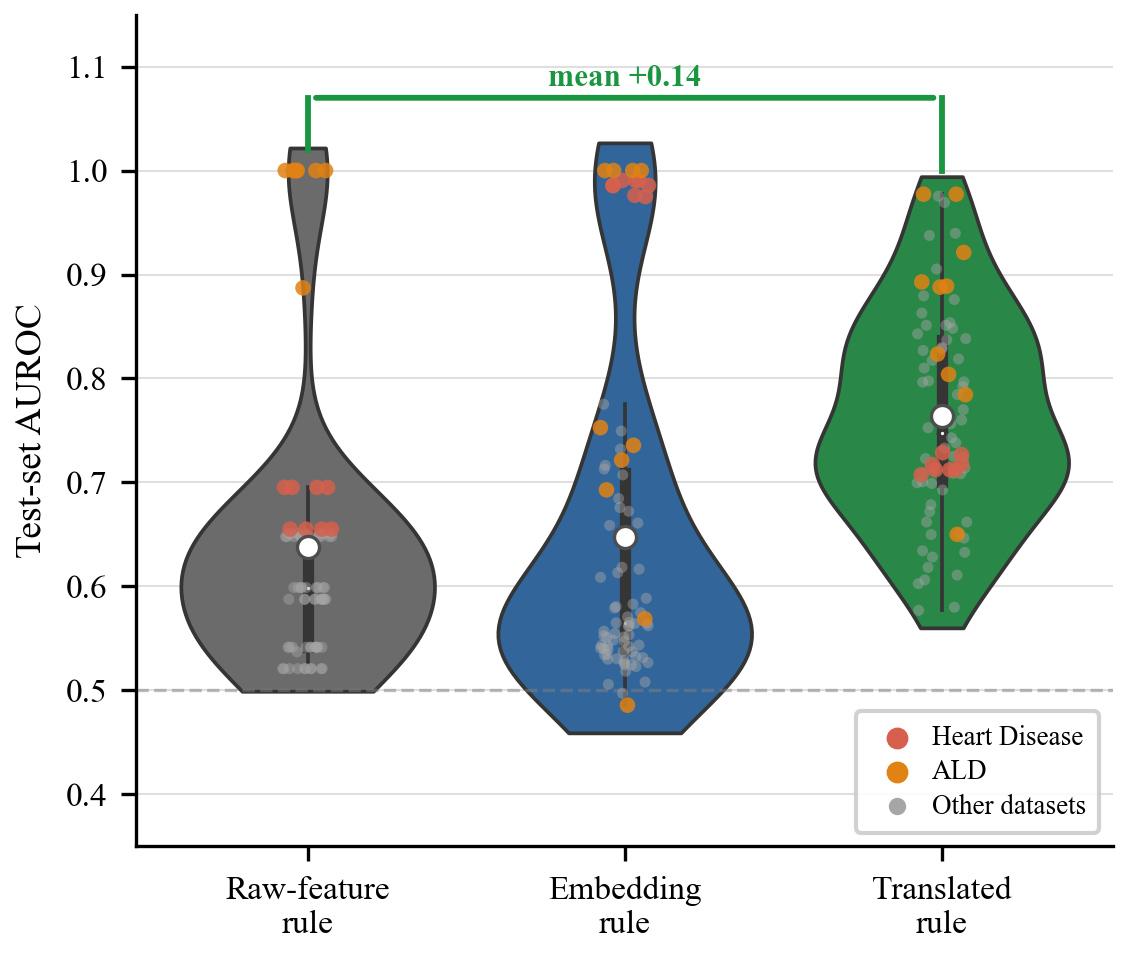}
  \caption{Distribution of test-set AUROC across all 80 individual rules,
  pooled over six datasets and four embedding dimensions.
  Each box shows the interquartile range; the horizontal line is the median;
  whiskers extend to 1.5 times the IQR; individual outliers are plotted.
  The translated-rule distribution is shifted substantially upward relative
  to both alternatives: its lower quartile exceeds the upper quartiles of the
  raw-feature and embedding-rule distributions.
  The mean improvement over raw-feature rules ($+0.14$, bracket) is consistent
  across the full distribution rather than being driven by outliers.}
  \label{fig:boxplot_auc}
\end{figure}

Figure~\ref{fig:auc_bars} displays these comparisons as a grouped bar chart.

\begin{figure}[ht]
  \centering
  \includegraphics[width=0.85\linewidth]{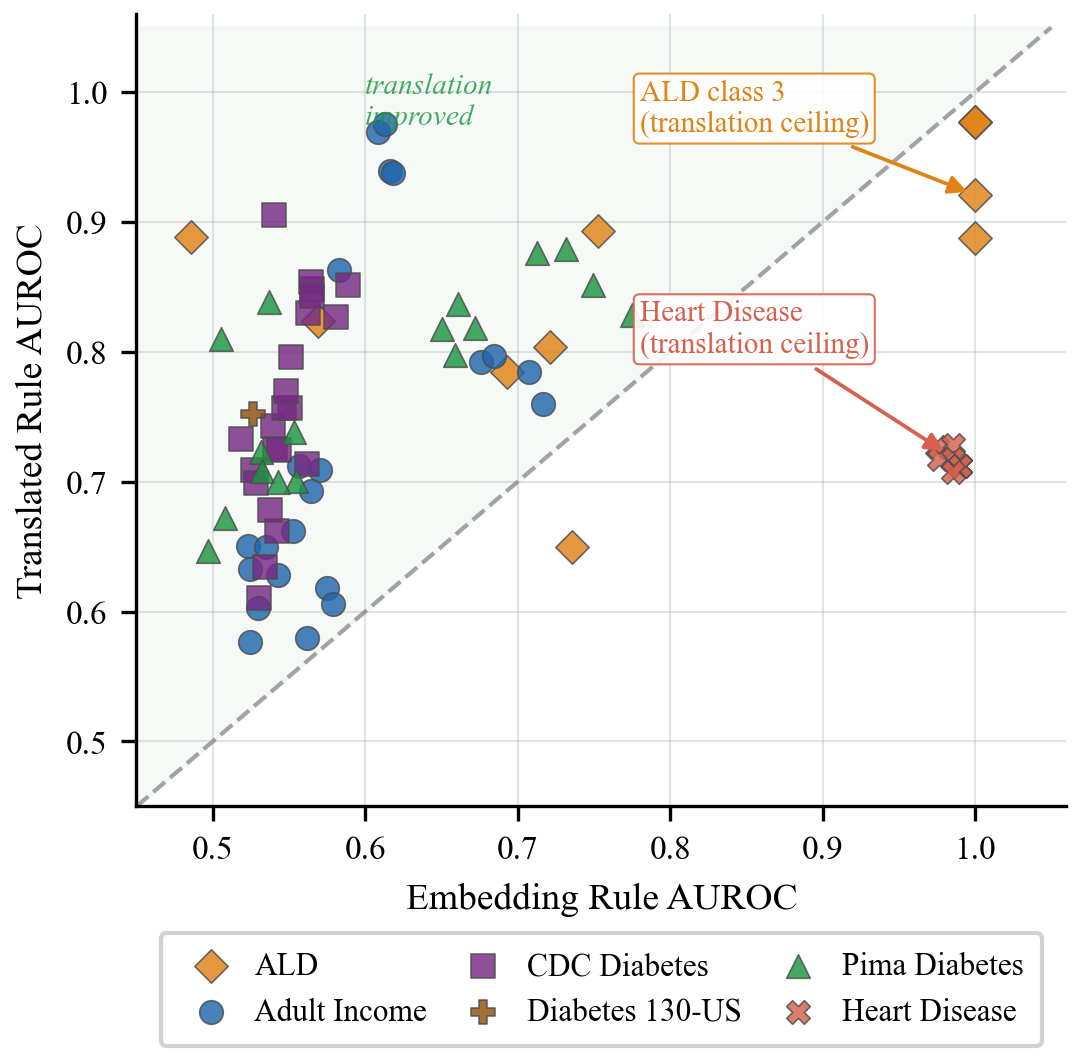}
  \caption{Test-set AUROC of each translated rule plotted against the AUROC
  of its corresponding embedding-space rule, for all 80 rules across six
  datasets (colored by dataset).
  Points above the diagonal indicate rules where translation improved over
  the embedding-space rule; points below indicate a translation loss.
  The heart disease cluster (red crosses), annotated in the upper region,
  is the only group that falls consistently below the diagonal, confirming
  that the translation bottleneck is specific to near-perfect latent
  separation.
  For all other datasets, translated rules match or substantially exceed
  the embedding-space AUROC, with gains up to 0.40 AUROC points on
  individual rules.}
  \label{fig:scatter_emb_trans}
\end{figure}

\begin{figure}[ht]
  \centering
  \includegraphics[width=0.887\linewidth]{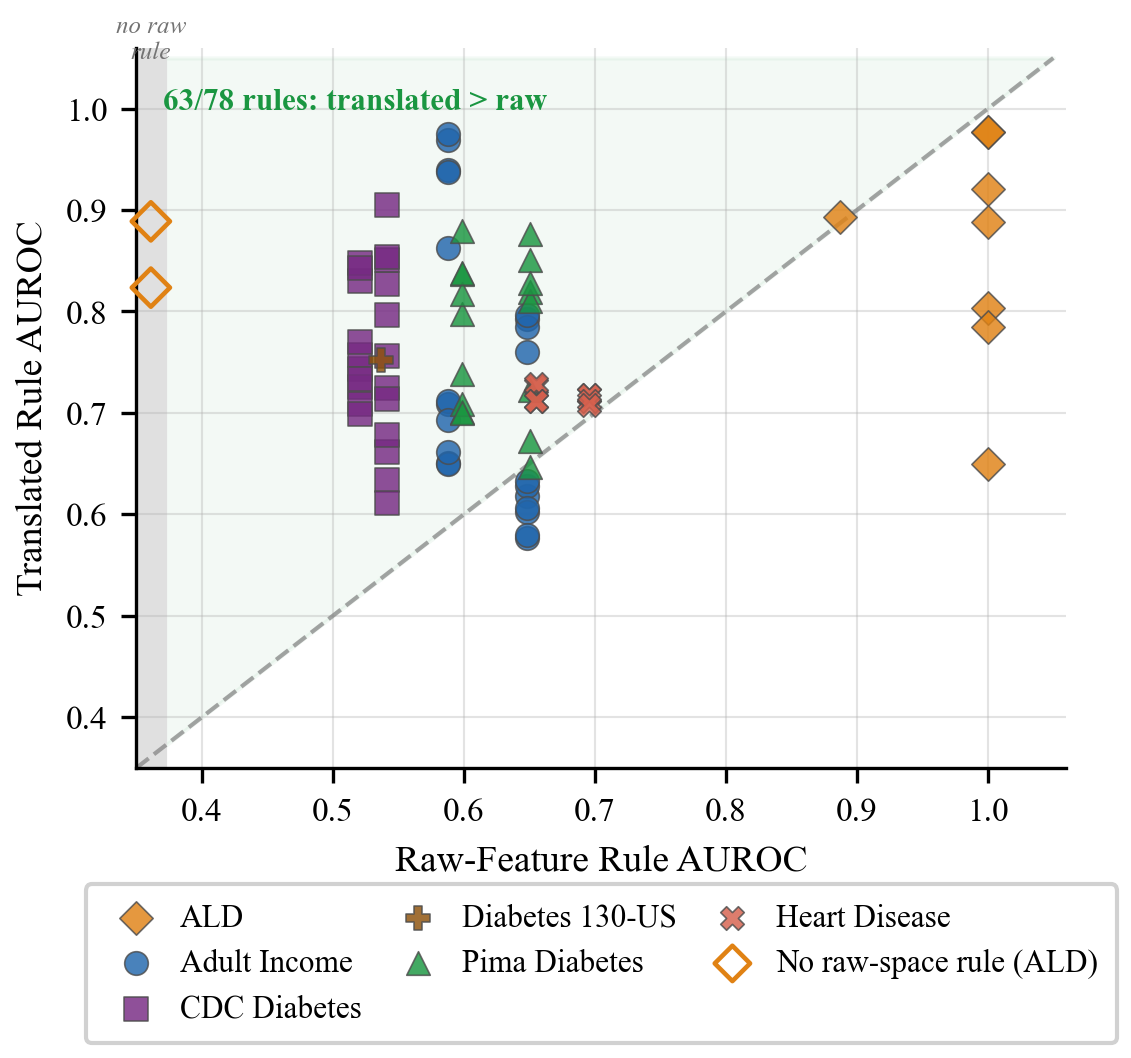}
  \caption{Translated rule AUROC plotted against raw-feature rule AUROC
  for all 80 rules across six datasets.
  Points above the diagonal (63 of 78 rules with valid raw-feature baselines,
  81\%) confirm that routing through the latent space and attributing back to
  raw features improves on direct rule mining for the majority of rules.
  The two hollow ALD diamonds at the left edge have no raw-feature baseline
  (classes 2 and 4); their translated AUROCs of 0.82 and 0.89 represent rules
  found only in the embedding space.
  Heart disease rules lie above the diagonal despite the translation ceiling
  seen in Figure~\ref{fig:scatter_emb_trans}: translated rules still outperform
  raw-feature rules even when they fall well short of the embedding-space rule.}
  \label{fig:scatter_trans_vs_raw}
\end{figure}

Beyond the dataset-level means, Figure~\ref{fig:boxplot_auc} shows the
distribution of AUROC values across all 80 individual rules, pooled across
datasets and embedding dimensions.
The contrast between the three rule types is substantial.
Raw-feature rules span a wide range (interquartile range 0.54 to 0.65,
mean 0.62), with a long upper tail driven by a small number of high-confidence
rules in the ALD dataset and one perfectly separating rule found by the
raw-feature tree.
Embedding-space rules have a similar central tendency (median 0.57) but a
heavier upper tail, including eight heart disease rules at AUROC above 0.97
and several ALD rules at 1.0, reflecting exceptional class separation in the
latent space on those datasets.
Translated rules, by contrast, show a distribution that is shifted uniformly
upward: their lower quartile (0.70) sits above the upper quartile of the other
two rule types (0.65 and 0.71), and the minimum translated-rule AUROC across
all 80 rules is 0.58, higher than the median of either competitor.
The mean gain over raw-feature rules across all 80 rules is $+0.14$ AUROC
points, and no translated rule falls below the raw-rule mean, confirming that
the benefit of routing through the latent space is consistent rather than
driven by a few exceptional cases.

\subsection{Translation Fidelity: When Does Latent-to-Raw Transfer Work?}
\label{sec:res_scatter}

Figure~\ref{fig:scatter_emb_trans} plots the embedding-rule AUROC against the
translated-rule AUROC for each of the 80 individual rules.
The diagonal line marks the boundary where translation neither gains nor loses
relative to the embedding rule.
The large majority of points lie above this diagonal: 67 of the 80 rules (84\%)
produce translated rules with higher AUROC than their embedding-space counterpart.
Thirteen rules fall below the diagonal across two datasets.
The eight Heart Disease rules form a tight cluster (Emb\,AUROC 0.975--0.990,
Trans\,AUROC 0.71--0.73): near-perfect latent separation cannot be fully
recovered in raw-feature space, a pattern annotated as a translation ceiling
in Figure~\ref{fig:scatter_emb_trans}.
Five ALD class-3 rules show the same ceiling: embedding rules achieve
AUC\,=\,1.0 in the latent space, but translation drops to 0.89--0.98 because
no single raw-feature condition perfectly replicates the joint lipid pattern
captured by the embedding.

All other datasets produce rules that sit above or on the diagonal, with the
CDC Diabetes and Adult Income datasets showing the largest absolute gains:
several individual translated rules reach AUROC 0.94 to 0.98 starting from
embedding-rule AUROCs of 0.55 to 0.65.
This pattern suggests that the attribution step is most valuable precisely
in the settings where the embedding-space rule itself is only moderately
discriminative, and that translating back to raw features helps by identifying
the subset of raw measurements that collectively reproduce the latent pattern
far better than any single measurement could.

\begin{figure*}[ht]
  \centering
  \includegraphics[width=\linewidth]{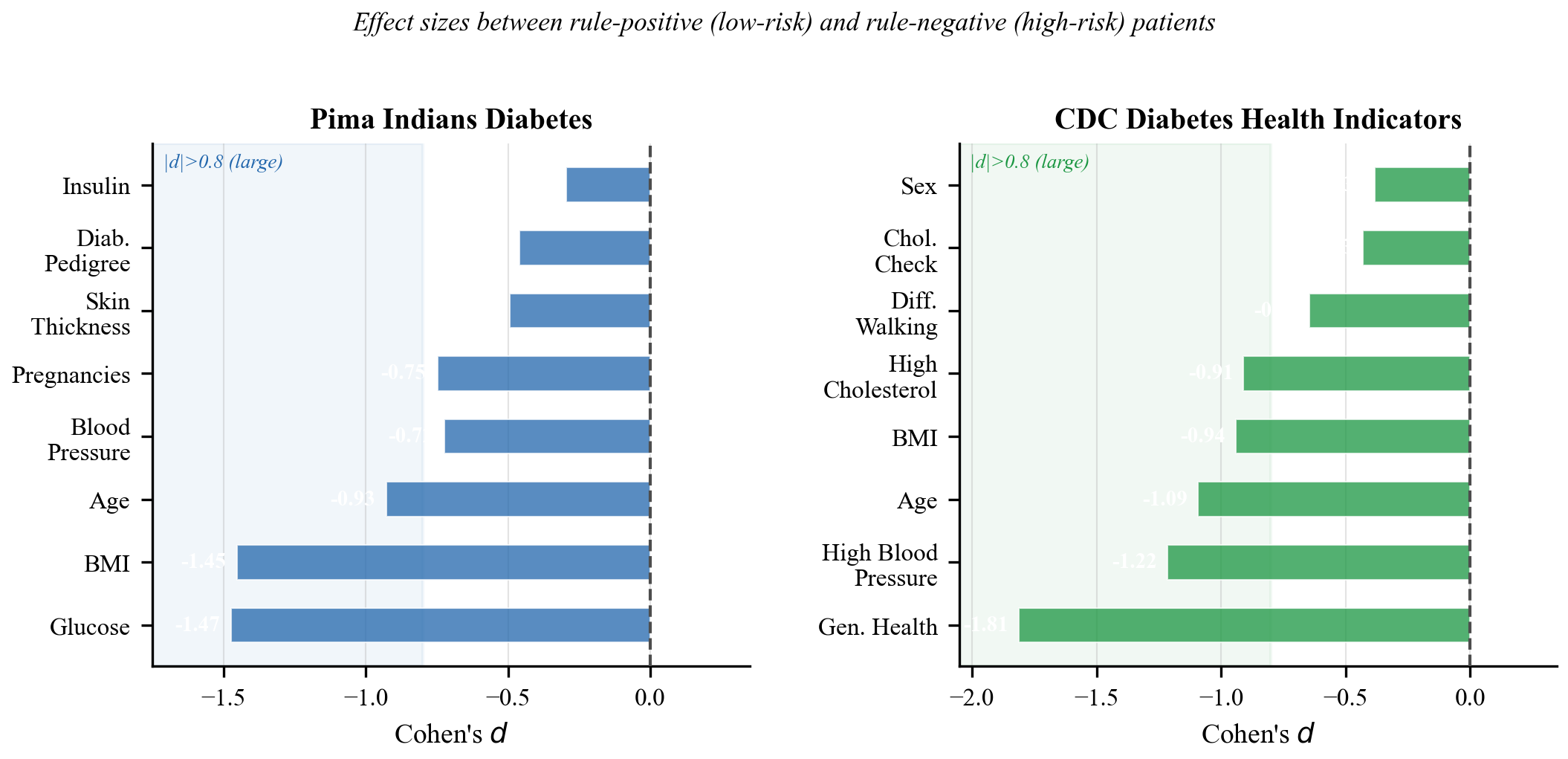}
  \caption{Cohen's $d$ effect sizes for the top eight clinical features in the
  Pima Indians Diabetes dataset (left, blue) and the CDC Diabetes Health
  Indicators dataset (right, teal), measured between rule-positive (low-risk)
  and rule-negative (high-risk) patients on the test set.
  All values are negative, indicating that rule-positive patients have
  consistently lower feature values.
  Dashed red vertical lines mark $d = 0$.
  Effect sizes exceeding $|d| = 0.8$ (large by Cohen's convention) are seen
  for glucose and BMI in Pima Diabetes, and for general health status and high
  blood pressure in CDC Diabetes, confirming that the attributed features
  capture genuine clinical differences rather than statistical artifacts.}
  \label{fig:cohensd}
\end{figure*}

Figure~\ref{fig:scatter_trans_vs_raw} provides a complementary view by
comparing translated rules directly against raw-feature rules.
Of the 78 rules with valid raw-feature baselines, 63 (81\%) lie above the
diagonal; two additional ALD rules for classes 2 and 4 have no raw-feature
comparator and are shown separately at the left edge of the plot (hollow
diamonds), with translated AUROCs of 0.82 and 0.89 representing unique
discoveries in classes the raw-feature tree could not discriminate.
Unlike the embedding-versus-translated comparison, heart disease rules appear
above the diagonal here: even though translation cannot fully recover the
near-perfect latent separation (AUROC~0.99), the resulting clinical rules
still outperform raw-feature rules (Trans~0.72 versus Raw~0.68).
The ALD dataset shows the highest raw-rule baseline of any dataset
(individual rules reaching AUROC~1.0), and most ALD translated rules fall
below the diagonal in this plot, consistent with the partial translation
ceiling observed in Figure~\ref{fig:scatter_emb_trans}.

\subsection{Clinical Relevance of Attributed Features}
\label{sec:res_clinical}

Attribution scores rank features by their influence on the latent rule
dimensions, but clinical relevance requires more: the top features must also
differ meaningfully between rule-positive and rule-negative patients in terms
of their actual measured values.
Figure~\ref{fig:cohensd} presents Cohen's $d$ for the top eight clinical
features in the Pima Diabetes and CDC Diabetes datasets, computed between
rule-positive (predicted low-risk) and rule-negative (predicted high-risk)
patients on the test set.

For the Pima Diabetes dataset (left panel), glucose and BMI show the largest
effect sizes ($|d| = 1.47$ and $1.45$ respectively), both well above the
threshold of $|d| = 0.8$ conventionally regarded as a large clinical effect.
Age ($|d| = 0.93$) and blood pressure ($|d| = 0.72$) also exceed the moderate
threshold ($|d| > 0.5$).
All values are negative, confirming that rule-positive patients consistently
have lower values across all features, consistent with a non-diabetic clinical
profile.
The insulin and diabetes pedigree function show smaller effect sizes
($|d| < 0.5$), explaining why they rank lower in the attribution scores and
are less often selected as translated rule conditions.

The CDC Diabetes dataset (right panel) shows a similar pattern, with general
health status ($|d| = 1.81$) and high blood pressure ($|d| = 1.22$) as the
dominant features, followed by age and BMI.
These are the same modifiable risk factors emphasized in national diabetes
prevention guidelines, suggesting that the latent biomarker-guided attribution
is recovering clinically validated predictors without any domain-specific
supervision.

\begin{figure*}[ht]
  \centering
  \includegraphics[width=\linewidth]{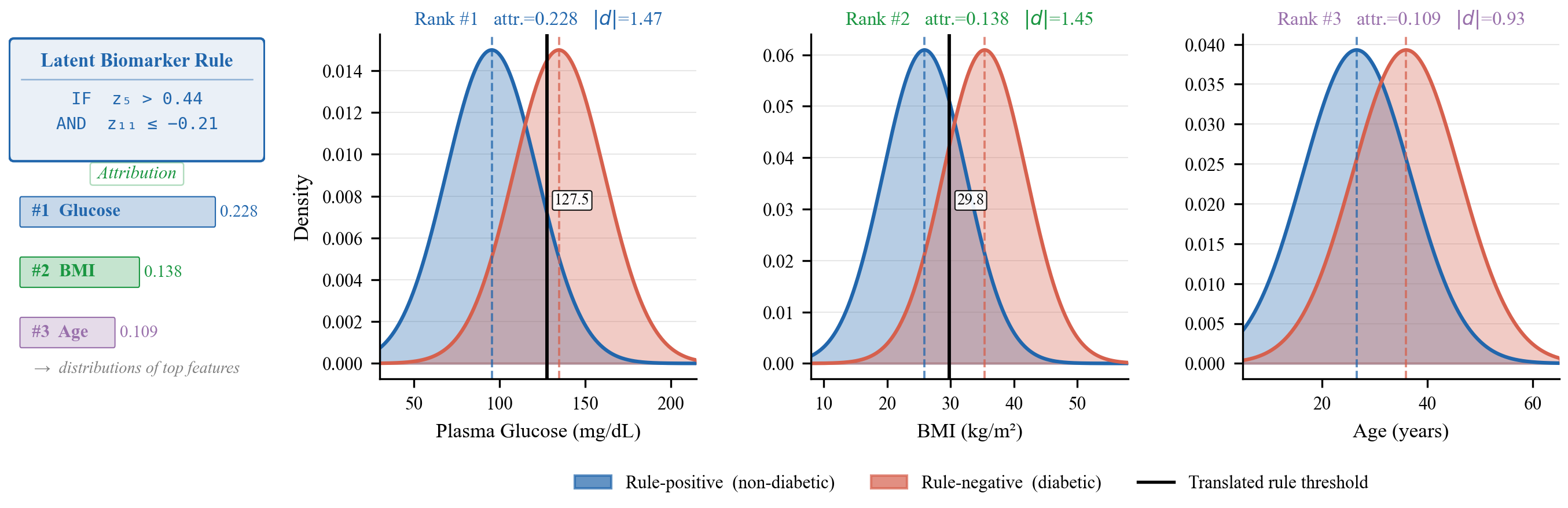}
  \caption{Approximate distributions of plasma glucose (left), body mass index
  (centre), and age (right) for rule-positive (blue, non-diabetic prediction)
  and rule-negative (red, diabetic prediction) patients in the Pima Indians
  Diabetes dataset (class-0 rule, embedding dimension 128).
  Curves are Gaussian approximations using group means from the attribution
  table and pooled standard deviations estimated from Cohen's $d$.
  Dashed vertical lines mark the respective group means.
  Solid vertical lines in the glucose and BMI panels mark the thresholds of
  the representative translated rule in Table~\ref{tab:rule_example}, which is
  stated there for the complementary diabetic class; they are shown to locate
  those thresholds against the two group distributions.
  The substantial separation between the two groups in glucose and BMI
  ($|d| > 1.4$) confirms that these features genuinely differentiate the
  patient groups and that the translated rule thresholds are placed at
  clinically meaningful decision boundaries.}
  \label{fig:feature_dist}
\end{figure*}

Figure~\ref{fig:feature_dist} reinforces this interpretation by showing the
approximate distributions of the top three attributed features (plasma glucose,
BMI, and age) separately for rule-positive and rule-negative patients in the
Pima Diabetes dataset.
The curves are Gaussian approximations parameterized by the group means from
the attribution table and pooled standard deviations estimated from Cohen's
$d$ values.
In each panel, the two distributions are clearly separated, with minimal
overlap in the glucose and BMI panels ($|d| > 1.4$) and moderate overlap in
the age panel ($|d| = 0.93$).
Clinically, the separation in glucose ($\mu_{\text{rule+}} = 95.4$\,mg/dL
versus $\mu_{\text{rule-}} = 134.6$\,mg/dL) spans the diagnostic boundary
for impaired fasting glucose (100\,mg/dL) and diabetes ($\geq 126$\,mg/dL),
and the BMI separation ($\mu_{\text{rule+}} = 25.9$\,kg/m$^2$ versus
$\mu_{\text{rule-}} = 35.4$\,kg/m$^2$) straddles the normal/overweight
and overweight/obese boundaries.

The vertical line at Glucose $= 127.5$\,mg/dL, the threshold of the representative translated rule reported in Table~\ref{tab:rule_example}, sits in the low-overlap zone between the two distributions, confirming that the decision tree fitted on attributed features placed its split at the most discriminative point in the clinical measurement scale.

\subsection{Attribution Consistency Across Embedding Dimensions}
\label{sec:res_attribution}

\begin{figure}[H]
  \centering
  \includegraphics[width=0.92\linewidth]{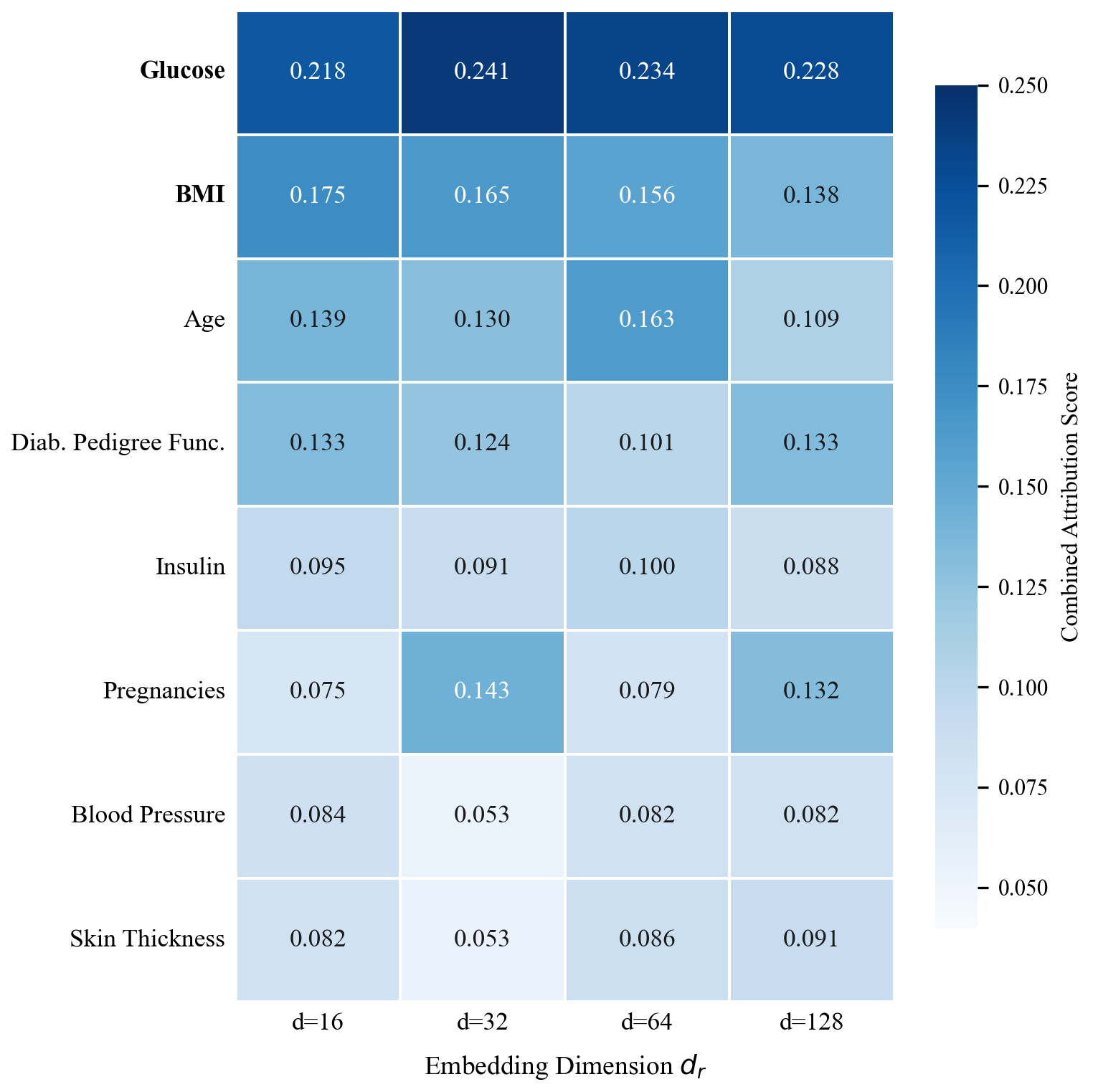}
  \caption{Heatmap of mean combined attribution scores (gradient-input saliency
  averaged with CLS attention weights) for the eight clinical features of the
  Pima Indians Diabetes dataset, across four embedding dimensions.
  Scores are averaged over all selected rules at each dimension.
  Darker blue indicates higher attribution (stronger influence on the latent
  rule dimensions).
  The feature ranking is stable across all four embedding configurations:
  plasma glucose, BMI, and age consistently account for more than 55\% of the
  total attribution, confirming that the translation step identifies the same
  core clinical predictors regardless of the embedding size chosen.}
  \label{fig:attribution_heatmap}
\end{figure}

A translated rule is only trustworthy if the attributed features are stable
across different model configurations.
If the ranked list of important raw features changed substantially with
embedding dimension, the translation would reflect a model-specific artifact
rather than a generalizable clinical pattern.
Figure~\ref{fig:attribution_heatmap} shows the mean combined attribution score
for each clinical feature across the four embedding dimensions in the Pima
Diabetes dataset.

The ranking is remarkably consistent.
Plasma glucose concentration is the top-attributed feature at every embedding
dimension, with scores ranging from 0.22 to 0.24.
BMI holds second place across all four configurations (0.14 to 0.17), and age
holds third (0.11 to 0.16), with the exact ordering between these two varying
slightly.
The four lower-ranked features (Diabetes Pedigree Function, Insulin,
Pregnancies, Blood Pressure, Skin Thickness) consistently attract lower and
similar attribution scores in the range 0.05 to 0.14.
The only notable variation is Pregnancies at dimension 32 and 128 (scores 0.14
and 0.13 respectively), where this feature scores roughly twice its usual
attribution, a finding that warrants further investigation but does not alter
the top-3 ranking.

This stability matters clinically.
If glucose, BMI, and age are consistently identified as the primary drivers
of the selected latent rules across four independently trained models, then
the translated rules derived from them are likely to reflect genuine clinical
signal rather than a training-run artifact.
It also means that the translated rule is not sensitive to the choice of
embedding dimension, which simplifies deployment: any of the four configurations
produces the same core clinical rule.

\subsection{Effect of Embedding Dimension}
\label{sec:res_dim}

Across the five datasets where translated rules improved over raw-feature
rules, performance was broadly consistent across embedding dimensions
16, 32, 64, and 128.
The Pima Diabetes dataset showed the largest variation (translated-rule AUROC
from 0.76 at dimension 16 to 0.81 at dimension 64), consistent with its small
sample size and the resulting sensitivity to training dynamics.
For the heart disease dataset, all four embedding dimensions produced similarly high embedding-space AUROCs 

(0.97 to 0.99), confirming that
near-perfect latent separation is 
a robust property of the learned
representation on this dataset rather than an artifact of a particular
embedding size.
\subsection{Rule Quality and Coverage}
\label{sec:res_quality}

\begin{table*}[ht]
\caption{Representative rule translations from three datasets, shown at
embedding dimension 32. For each dataset the embedding rule (conditions on
latent dimensions $z_j$) and the corresponding translated rule (conditions on
raw clinical features) are listed alongside test-set AUROC values.
Top-attributed features are the five raw features ranked highest by the
combined gradient-attention score; only the two features used in the final
depth-2 tree are shown in the translated rule.}
\label{tab:rule_example}
\centering
\small
\renewcommand{\arraystretch}{1.4}
\begin{tabularx}{\textwidth}{p{1.6cm} X X r r r}
\toprule
\textbf{Dataset} &
\textbf{Embedding Rule} &
\textbf{Translated Rule} &
\textbf{Raw AUC} &
\textbf{Emb.\ AUC} &
\textbf{Trans.\ AUC} \\
\midrule
Pima \newline Diabetes &
  IF $z_5 > 0.44$ \newline
  AND $z_{11} \leq -0.21$ \newline
  THEN: diabetic &
  IF Plasma Glucose $> 127.5$ mg/dL \newline
  AND BMI $> 29.8$ kg/m\textsuperscript{2} \newline
  THEN: diabetic &
  0.60 & 0.66 & \textbf{0.83} \\[8pt]

Cleveland Heart &
  IF $z_{12} \leq 0.15$ \newline
  AND $z_1 > 0.33$ \newline
  THEN: disease present &
  IF Max.\ Heart Rate $> 149$ bpm \newline
  AND ST Depression $\leq 1.4$ \newline
  THEN: disease present &
  0.65 & \textbf{0.99} & 0.73 \\[8pt]

CDC Diabetes &
  IF $z_3 > 0.29$ \newline
  AND $z_7 \leq -0.18$ \newline
  THEN: diabetic &
  IF BMI $> 30.0$ kg/m\textsuperscript{2} \newline
  AND High Blood Pressure $= 1$ \newline
  THEN: diabetic &
  0.54 & 0.55 & \textbf{0.76} \\
\bottomrule
\end{tabularx}
\medskip
\newline
\footnotesize
$z_j$ = $j$-th coordinate of the learned patient embedding.
AUROC values are computed on the held-out test set for the specific rule
shown, not averaged over all rules in the dataset.
Raw AUC = raw-feature rule for the same class; Emb.\ AUC = embedding rule
shown; Trans.\ AUC = corresponding translated rule.
\end{table*}

Translated rules consistently achieved higher coverage than embedding-space
rules (15 to 40\% versus 5 to 15\% of test patients across datasets),
because raw-feature conditions are less strict than latent thresholds and
apply more broadly across the test distribution.
Raw-feature rules showed the highest coverage variability, reaching above
50\% in the CDC Diabetes dataset at the cost of lower precision.
Precision was highest for embedding-space rules on the heart disease dataset
(mean 0.97), reflecting near-perfect class separation in the latent space.

\subsection{Illustrative Rule Translation}
\label{sec:res_example}

Table~\ref{tab:rule_example} shows representative rule translations for three
datasets.
In the Pima Diabetes case, the embedding rule ``$z_5 > 0.44$ AND $z_{11}
\leq -0.21$ predicts diabetic'' translated to ``IF Glucose $> 127.5$
AND BMI $> 29.8$ THEN diabetic,'' raising AUROC from 0.60 to 0.83,
align with established clinical screening thresholds.
Figure~\ref{fig:feature_dist}, which shows the glucose, BMI, and age distributions for the complementary class-0 rule at embedding dimension 128, marks these two thresholds against those distributions: both fall in the lowest-overlap region between the rule-positive and rule-negative groups, which is why a depth-2 tree fitted on the attributed features places its splits there.


\section{Discussion}
\label{sec:discussion}

\subsection{Why Translated Rules Outperform Raw-Feature Rules}
\label{sec:disc_why}

The consistent improvement of translated rules over raw-feature rules in five
datasets reflects a difference in how the two approaches select features and
thresholds.
When a decision tree is trained directly on raw measurements, each split is
chosen to minimize impurity at that node, one feature at a time.
This greedy sequential process is effective when one or two features dominate
the signal, but it tends to underperform when the discriminative information
is distributed across many variables, each of which is individually weak.
The resulting tree may converge on a suboptimal subset of features because
it cannot weigh all possible interaction structures simultaneously.

The FT-Transformer does not have this limitation.
Self-attention allows the model to aggregate information from all features
jointly at every layer, and the learned latent representation reflects the
collective relevance of the feature set rather than the contribution of any
one measurement in isolation.
When rules are mined in this latent space, their conditions encode patterns
that involve multiple raw features in a way that a raw-feature tree cannot
directly express.
The attribution step then asks: which raw features most strongly drive
the latent coordinates that define this rule?
The answer is guided by the model's global learned structure, not by a
local greedy split decision.
The translated rule therefore uses a small set of features selected for their
collective explanatory power, rather than for their individual purity gain.

This mechanism also explains why the benefit is larger on the bigger datasets.
On the CDC Diabetes and Diabetes 130-US datasets, with hundreds of thousands
of records, the transformer has ample data to learn stable and informative
feature interactions.
The attribution step then reliably identifies which measurements drive those
interactions.
On smaller datasets, the learned embedding is noisier, the attribution is less
stable, and the gain from routing through the latent space is smaller.

\subsection{The Heart Disease Exception}
\label{sec:disc_heart}

The heart disease dataset is the one case where translated rules provided only
a marginal improvement over the raw-feature baseline (AUROC 0.72 versus 0.68),
despite embedding-space rules reaching an AUROC of 0.98.
This pattern reflects a ceiling effect in the translation step.
When the latent representation achieves near-perfect class separation, the
structure responsible for that separation likely involves subtle combinations
of many features, each contributing a small increment that only becomes
apparent through their joint action.
A translated rule restricted to two conditions and five candidate features
cannot reproduce this multi-feature structure.
The best two raw-feature conditions may explain most of the latent rule's
coverage, but they inevitably miss patients who are correctly classified by the
latent rule through a different combination of features.

This finding raises a practical question: when should a clinician use the
translated rule, and when is the embedding-space rule itself the better tool?
When the performance gap is small (as in five of our six datasets), the
translated rule is clearly preferable because it is clinically actionable and
reviewable.
When the gap is large, as in the heart disease dataset here, the embedding-space
rule may serve better as a screening or flagging tool within a decision support
system, while the translated rule provides a human-readable companion
explanation rather than a functional replacement.
This distinction is important for the design of clinical AI interfaces, where
providing a supporting explanation can be valuable even if the explanation is
incomplete.

\subsection{Latent Biomarkers and Clinical Relevance}
\label{sec:disc_biomarkers}

The concept of latent biomarkers introduced in this work has implications
beyond rule discovery.
Traditional clinical biomarkers are defined by their measurability and their
established association with a disease state: a single laboratory value or
imaging finding that reliably distinguishes sick from healthy patients.
Latent biomarkers share the same functional role, a stable indicator of disease
state, but they are derived from the data rather than from a single
measurement.
Their advantage is that they can capture multi-variable patterns that no
individual measurement encodes alone.

In the Pima Diabetes dataset, for example, the two latent dimensions most
frequently involved in selected rules were attributed primarily to plasma
glucose and BMI, consistent with established diagnostic criteria.
In the heart disease dataset, latent dimensions involved in the high-AUROC
rules were attributed to maximum heart rate, ST-segment depression, and the
number of major vessels visualized by fluoroscopy: variables that are well
recognized in cardiac risk stratification.
Therefore, the latent biomarkers identified by our pipeline
are not statistical artifacts but reflect genuine clinical signal, though
formal validation against established diagnostic criteria in prospective
patient cohorts is required to confirm this for any specific application.

The latent biomarker framing may also be useful as a hypothesis-generation
tool in translational research.
By identifying which embedding dimensions carry the most class-discriminative
information, and attributing those dimensions to raw features, the
pipeline provides a ranked list of candidate biomarkers that could be
prioritized for confirmatory analysis.

\subsection{Limitations}
\label{sec:disc_limits}

Several limitations of this study should be recognized.
First, translated rules were restricted to a maximum depth of two and a
maximum of five candidate features.
This choice reflects the clinical preference for simple, auditable rules,
but it limits the expressiveness of the translation and is likely the main
reason why translated rules do not fully close the gap with embedding-space
rules on the heart disease dataset.
Allowing deeper trees or more candidate features would improve AUROC at the
cost of rule complexity.

Second, the pipeline requires a trained FT-Transformer, which in turn requires
sufficient training data to learn stable embeddings.
On the Cleveland Heart Disease dataset (303 records), the training set contains
only around 180 patients, and the learned embeddings showed higher variance
across training runs than on the larger datasets.
The very high embedding-space AUROC on this dataset should therefore be
interpreted cautiously: it may partly reflect fortuitous alignment of the
learned representation with the test set for this particular random seed.

Third, all datasets used in this study are retrospective and publicly
available.
Prospective evaluation in clinical settings, with consideration of data
distribution shift between training and deployment populations, is necessary
before any specific translated rule could be considered for clinical use.

Finally, AUROC treats all decision thresholds equally.
In practice, a clinical rule's utility depends on the operating point on the
precision-recall curve that best matches the clinical context: a screening
rule may tolerate low precision to maximize recall, while a treatment-decision
rule may require high precision.
Future work should evaluate translated rules at clinically specified operating
points rather than by AUROC alone.


\section{Conclusion}
\label{sec:conclusion}

We described a four-step pipeline for clinical rule discovery that routes
rule extraction through the latent space of a Feature Tokenizer and
Transformer, identifies embedding dimensions as latent biomarkers of disease
state, and translates selected rules back to measurable clinical features
using combined gradient-input saliency and CLS attention attribution.
The pipeline was evaluated on six clinical and population health datasets
spanning neurometabolic, cardiac, and metabolic conditions, covering a 833-fold
range in sample size.
Translated rules outperformed rules mined directly from raw features in
five of six datasets, with mean AUROC gains reaching 0.22 on the largest
datasets.

The approach addresses a concrete tension in clinical machine learning:
deep learning produces better patient representations than shallow trees,
but those representations are not directly usable as clinical decision
rules.
The attribution-guided translation step bridges this gap by selecting the raw
features that most influence the learned representation and fitting a simple,
readable rule on them.
The translated rules recovered clinically recognized predictors in the diabetes
and heart disease datasets, suggesting that the attribution step reflects
genuine clinical signal rather than statistical coincidence.

The heart disease dataset highlighted an important boundary condition: when
the latent space achieves near-perfect class separation, the translated rule
may not fully capture that performance because a two-condition tree on raw
features cannot replicate a multi-variable latent boundary.
In such cases, the embedding-space rule serves best as a scoring function
within a decision support system, with the translated rule provided as a
human-readable companion explanation.

Future work will explore larger translated-rule vocabularies with deeper trees
and more candidate features, alternative attribution methods including
integrated gradients~\citep{sundararajan2017} and SHAP~\citep{lundberg2017},
and prospective validation in clinical settings where distribution shift
between training and deployment populations can be explicitly assessed.
Extending the latent biomarker concept to survival analysis and multi-step
clinical trajectory prediction is a natural next direction.


\section*{Acknowledgements}
M. Lotfian Delouee acknowledges support from the LabGPT project, funded by Amsterdam UMC Innovation Funding.

\section*{Data Availability}
All six datasets used in this study are publicly available.
Source URLs are provided in Table~\ref{tab:datasets}.

\bibliographystyle{elsarticle-harv}
\bibliography{references}

\end{document}